\documentclass[letterpaper,10pt,conference]{ieeeconf}  %
\IEEEoverridecommandlockouts 
\let\labelindent\relax
\usepackage{cite}
\usepackage{amsmath,amssymb,amsfonts}
\usepackage{algorithmic}
\usepackage{graphicx}
\usepackage{textcomp}
\usepackage{xcolor}
\usepackage[inline]{enumitem}

\makeatletter
\let\NAT@parse\undefined
\makeatother
\usepackage[breaklinks=true,hidelinks,bookmarks=false]{hyperref}

\usepackage[dvipsnames,table]{xcolor}
\usepackage{soul}
\usepackage[T1]{fontenc}  
\usepackage[cmintegrals]{newtxmath}
\usepackage{bm}
\usepackage{amsmath}

\usepackage{xspace} 
\usepackage{multirow} 

\usepackage{graphicx}
\graphicspath{{figures/}}

\usepackage{relsize}
\usepackage{nicematrix}

\usepackage{array}

\usepackage{makecell}

\usepackage{url}

\usepackage{array}
\usepackage{graphicx}
\usepackage{multirow}
\usepackage{amsmath}
\usepackage{amssymb}
\usepackage{tabularx}
\usepackage{booktabs}
\usepackage{enumerate}
\usepackage{array,multirow}
\usepackage{tablefootnote}
\usepackage{enumitem}
\usepackage[caption=false]{subfig}
\usepackage[export]{adjustbox}
\usepackage{pifont}
\usepackage[T1]{fontenc}
\usepackage{soul} %

\usepackage{flushend}
\usepackage{microtype}

\usepackage{dsfont}
\usepackage{etoolbox}

\DeclareMathOperator*{\argmin}{arg\,min}

\definecolor{exemplargreen}{HTML}{f3d6d7}
\definecolor{searchred}{HTML}{f3d6d7}
\definecolor{darkgreen}{HTML}{445b00}
\definecolor{lightergray}{HTML}{dddddd}
\definecolor{goodgreen}{rgb}{0.0, 0.56, 0.0}
\definecolor{badgray}{HTML}{666666}

\definecolor{buscasota}{rgb}{0.9, 0.95, 1.}

\definecolor{rankacolor}{HTML}{bebeff}
\definecolor{rankbcolor}{HTML}{d9d9ff}
\definecolor{rankccolor}{HTML}{e9e9ff}

\newcommand{\best}[1]{\textbf{#1}}

\newcommand{\method}{\mbox{PPT}\xspace}

\makeatletter
\DeclareRobustCommand\onedot{\futurelet\@let@token\@onedot}
\def\@onedot{\ifx\@let@token.\else.\null\fi\xspace}

\def\eg{{e.g}\onedot}
\def\ie{{i.e}\onedot} 
 
 \def\vs{{vs}\onedot}

\makeatother

\newif\ifshowedits

\newcommand{\addeditor}[3]{%
  \definecolor{#1color}{rgb}{#3}
  \expandafter\newcommand\csname #1\endcsname[1]{%
  \ifshowedits
    {\color{#1color} ##1}%
  \else
    {##1}%
  \fi
  }%
  \expandafter\newcommand\csname #1rmk\endcsname[1]{%
  \ifshowedits
    {\color{#1color} {\bf [#2: ##1]}}
  \fi
  }%
  \expandafter\newcommand\csname #1rpl\endcsname[2]{%
  \ifshowedits
    {\color{#1color} ##1 \sout{##2}}
  \else
    {##1}
  \fi
  }%
}

\usepackage{colortbl}
\newcolumntype{a}{>{\columncolor{blue!15}}c}

\let\titleold\title
\renewcommand{\title}[1]{\titleold{#1}\newcommand{\thetitle}{#1}}

\usepackage{wrapfig}

\addeditor{eloi}{EZ}{0.0, 0., 0.6}
\addeditor{THV}{THV}{0.0, 0.5, 0.9}
\addeditor{yuan}{YY}{0.95, 0.55, 0.15}
\addeditor{alex}{AB}{0.6, 0.4, 1.0}
\addeditor{xyh}{XYH}{0.0, 0.0, 0.0}
\addeditor{mat}{MC}{1.0, 0.3, 1.0}
\usepackage{cleveref}
\showeditstrue

\title{\LARGE \bf PPT: Pretraining with Pseudo-Labeled Trajectories\\ for Motion Forecasting}

\author{
Yihong Xu\textsuperscript{1*} 
\quad Yuan Yin\textsuperscript{1*} \quad Éloi Zablocki\textsuperscript{1} \quad 
{Tuan-Hung Vu}\textsuperscript{1} \quad {Alexandre Boulch}\textsuperscript{1} \quad {Matthieu Cord}\textsuperscript{1,2}
\thanks{$^{1}$ Valeo.ai, Paris, France; Email: \texttt{firstname.lastname@valeo.com}}
\thanks{$^{2}$ Sorbonne Université, CNRS, ISIR, F-75005 Paris, France}%
\thanks{Corresponding author: Y. Xu, \texttt{yihong.xu@valeo.com}}
\thanks{$^*$ Equal contribution.}
}

\begin{document}
\pagenumbering{arabic}
\maketitle
\pagestyle{plain}

\begin{abstract}
Accurately predicting how agents move in dynamic scenes is essential for safe autonomous driving. State-of-the-art motion forecasting models rely on datasets with manually annotated or post-processed trajectories. However, building these datasets is costly, generally manual, hard to scale, and lacks reproducibility. They also introduce domain gaps that limit generalization across environments.
We introduce \method{} (Pretraining with Pseudo-labeled Trajectories), a simple and scalable pretraining framework that uses unprocessed and diverse trajectories automatically generated from off-the-shelf 3D detectors and tracking. Unlike data annotation pipelines aiming for clean, single-label annotations, \method{} is a pretraining framework embracing off-the-shelf trajectories as useful signals for learning robust representations.
With optional finetuning on a small amount of labeled data, models pretrained with \method{} achieve strong performance across standard benchmarks, particularly in low-data regimes, and in cross-domain, end-to-end, and multi-class settings. \method{} is easy to implement and improves generalization in motion forecasting.
\end{abstract}

\section{Introduction}
\label{sec:intro}
Anticipating how a scene will evolve is a key requirement in human-robot interaction scenarios, such as assisted and autonomous driving. Forecasting the future motions of nearby agents is critical for safe trajectory planning and avoiding collisions in these settings.
\xyh{To support progress in motion forecasting (MF), datasets like nuScenes (NUS) \cite{nuscenes}, Waymo Open Dataset (WOD) \cite{waymo}, and Argoverse 2 (AV2) \cite{wilson2023argoverse} have become widely adopted. These datasets provide vectorized driving scenes with ground-truth agent trajectories, typically obtained through expensive manual annotation. In more recent motion forecasting datasets~\cite{womd, av2mf}, automatic pipelines have been proposed to achieve near-human-level annotations with lower cost, relying on extensive post-processing.} 

{
 \xyh{Training motion forecasting models \textit{solely}} on the human-annotated, post-processed datasets comes with significant limitations.
First, dataset creation \xyh{with human annotations} is expensive and labor-intensive, limiting scalability.
Second, the post-processing pipelines are often opaque and dataset-specific, involving hand-tuned post-processing steps that are difficult to reproduce or apply to new datasets. \xyh{Also, the selection of a single trajectory for each agent in these pipelines removes the data diversity.}
Third, differences in building datasets can introduce domain gaps: models trained on one dataset often perform poorly on others~\cite{feng2024unitraj, xu2024towards}.
}

{
To address these challenges, we introduce \method{}---Pretraining with Pseudo-labeled Trajectories---a simple and scalable \xyh{pretraining framework} for motion forecasting. \xyh{Existing motion forecasting pretraining strategies~\cite{xu2022pretram, cheng2023forecast, bhattacharyya2023ssl} heavily rely on annotated data and mostly pretrain only the output features of the encoder through masking.} Instead, \method{} pretrains models on pseudo-labeled data, followed by optional finetuning on a smaller amount of annotated ground truth. 
\method{} uses a fully automatic pipeline to generate pseudo-labeled trajectories. 
We apply off-the-shelf 3D object detectors~\cite{wang2023dsvt,shi2023pv,chen2023voxelnext, liu2023sparsebev, YinSC0YFW24, YinZK21} to estimate agent positions in individual frames, and then apply a lightweight, non-learning tracker~\cite{Weng2020_AB3DMOT, YinZK21} to associate these detections over time.

The resulting pseudo-labeled trajectories exhibit two key properties: they are noisy and diverse.
Indeed, compared to human annotations, pseudo-labeled trajectories are less precise, reflecting the realistic imperfections of detector outputs and simple tracking.
Moreover, instead of producing a single `ground-truth' trajectory per agent, we aggregate multiple pseudo-trajectories from different detectors.
\xyh{Fundamentally different from existing automatic pipelines with post-processing~\cite{womd, av2mf}, \method{} \emph{does not} aim to produce 'perfect' and a unique annotation per agent that emulates human annotators. \method{} is \textit{a simple pretraining framework} that embraces noise and diversity (from different trackers).} While imperfections in trajectories are usually viewed as weaknesses, \xyh{we show that post-processing is unnecessary and off-the-shelf tracking trajectories are surprisingly useful for pretraining.} The real-world variability encourages models to learn robust and generalizable representations.
}

{
A major advantage of \method{} is its flexibility. Because our pseudo-labeling pipeline is fully automated and dataset-agnostic, we can combine data from multiple sources without manual harmonization. This enables large-scale pretraining that enhances generalization—a trend also observed in recent works~\cite{feng2024unitraj}.
Moreover, while most motion forecasting models are trained from scratch, \method{} provides a valuable initialization. It allows models to first learn general forecasting dynamics from noisy and unlabeled data, and then specialize to any target domain with limited labeled data.
}

\xyh{In summary, we introduce \method{}, a simple and novel pretraining approach for motion forecasting. To our best knowledge, \method{} is the first pretraining strategy for motion forecasting models leveraging diverse pseudo-labeled trajectories from multiple off-the-shelf trackers. Some insightful observations are discovered:}
\begin{itemize}[noitemsep, topsep=0pt, left=5pt]

    \item \method{} significantly reduces the need for human annotations, and is especially effective in annotation-efficient regimes, where only 1$\sim$10\% of the ground-truth labeled trajectories are used for finetuning.
    \item \xyh{Both the inherent noise of pseudo-labeled trajectories and their variability contribute to the performance gain with PPT pretraining.}
    \item \xyh{We demonstrate that post-processing is not necessary and HD maps is optional for motion forecasting pretraining.}
    \item \method{} is widely applicable in various motion forecasting settings. It improves generalization across domains, including strong performance on out-of-distribution scenarios, end-to-end, and multi-class motion forecasting settings.
\end{itemize}

\section{Related Work}
\label{sec:related-work}
\noindent\textbf{Motion forecasting\;} consists of predicting the future trajectory for the agents of interest (\eg vehicles or 10 different classes in AV2 MF~\cite{av2mf}) based on their past trajectories, agent interactions, and map information. Conventional motion prediction methods~\cite{salzmann2020trajectron++, kim2021lapred, mtr, cui2023gorela, Wayformer, covernet, yuan2021agentformer, densetnt, multipath, multipath++, benyounes2022cab, mfp, mtp, mosa, wang2023ganet, xu2024annealed} focus on architectural design to improve trajectory prediction and assume that past trajectories are obtained from \xyh{human-labeled or post-processed} datasets~\cite{waymo, nuscenes, wilson2023argoverse}.
In contrast, end-to-end forecasting methods~\cite{xu2024towards, hu2023planning, gu2023vip3d, xu2024valeo4cast} focus on the real-world deployment condition, where past trajectories are estimated from a perception pipeline, consisting of detection and tracking. However, as noted in~\cite{xu2024towards}, current end-to-end \textit{trained} models~\cite{gu2023vip3d, hu2023planning} exhibit lower performance compared to modular approaches~\cite{xu2024valeo4cast}, and effective strategies to address perception model imperfections remain unclear. In this work, we aim to propose a simple {and} powerful pretraining framework that improves both paradigms with diverse pseudo-labeled trajectories and minimum annotation cost.

\noindent\textbf{Motion forecasting pretraining.\;}
Current motion forecasting heavily relies on the accurate annotations provided by the datasets and trains the models from scratch. Recently, several self-supervised methods for motion forecasting have emerged~\cite{xu2022pretram, cheng2023forecast, bhattacharyya2023ssl}. Specifically, in \cite{xu2022pretram}, Xu et al. pretrain map and trajectory encoders through contrastive learning, encouraging feature similarity between positive trajectory-map pairs. Authors in~\cite{bhattacharyya2023ssl} define several pretext tasks, such as lane masking, lane node distance prediction, and maneuver classification of agents of interest. Cheng et al. in \cite{cheng2023forecast} mask the lane and past/future trajectories and train a network to reconstruct the masked values. 
Notably, these pretraining methods \textit{still rely on human annotations} and generally improve features of a sub-module (encoders) through contrastive learning or masking. In \cite{yang2024visual}, Yang et al. pretrain a perception backbone using images for various downstream tasks, but focus only on single-dataset settings. %
Instead, we position our proposed strategy as \emph{task-specific pretraining}, which performs pretraining using pseudo-labeled data with the same optimization objective of motion forecasting.
Our goal is to explore diverse off-the-shelf trajectories for forecasting pretraining without directly using any annotated trajectories.

\noindent\textbf{Pseudo-labeling.\;}
Traditionally, driving datasets~\cite{wilson2023argoverse,nuscenes,waymo} rely heavily on human annotation processes to generate agent trajectory labels, a costly approach that limits scalability.
To address this, recent motion forecasting datasets such as WOMD~\cite{womd} and AV2 MF~\cite{av2mf} use automatic annotation pipelines coupled with complex post-processing steps to generate high-quality, near-human single-label trajectories.  
\xyh{PPT has a different goal, which is to propose a simple pretraining framework, for motion forecasting, without relying on accurate annotations.} Therefore,~\method{} embraces the raw tracking trajectories with various camera-~\cite{wang2022detr3d,liu2023sparsebev}, LiDAR-~\cite{YinZK21,wang2023dsvt,shi2023pv,chen2023voxelnext} or LiDAR-camera-based~\cite{LiuTAYMRH23,YinSC0YFW24} 3D detectors and a \textit{non-learnable} tracker based on geometry cues~\cite{Weng2020_AB3DMOT,YinZK21}. We note that the pseudo-labeling in~\method{} differs fundamentally from WOMD and AV2 MF in three key regards:~(1)~we do not aim to achieve clean labels, but embrace noisy and diverse trajectories for MF pretraining.~(2)~No post-processing like track smoothing~\cite{av2mf, womd} is performed to reduce noise.~(3)~No manual selection of a single trajectory per agent discarding the trajectory diversity, which we show important for pretraining. (4) Moreover, with diverse pseudo labels,~\method{} exhibits higher generalization ability across domains (\autoref{tab:cross_dataset_evaluation}).

\section{\method{} Framework}
\label{sec:method}

\subsection{Conventional Training for Motion Forecasting}
\begin{figure*}[t]%
    \centering
    \includegraphics[width=0.8\linewidth]{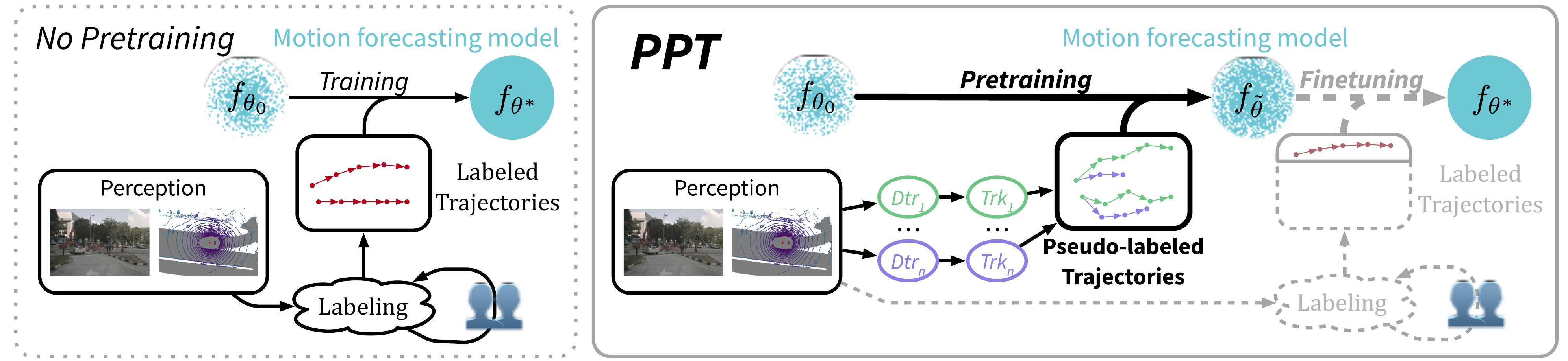}
  \caption{\textbf{Illustration of PPT.} On the left, we show the conventional approach to training a motion forecasting model from scratch with annotated labels, often from human labeling. By contrast, on the right, we present a pretraining pipeline with pseudo-labeled trajectories from different 3D detectors and (non-learning) trackers ($\mathit{Dtr}, \mathit{Trk}$), followed by an optional finetuning phase (shaded in \textcolor{gray}{gray}).
  }
    \label{fig:ppt}
\end{figure*}
Motion forecasting models are typically trained on labeled datasets composed of human-annotated trajectories within a specific domain.
We denote such a dataset as $\mathcal{T} = \{\tau_i\}_{i}$, where $\tau_i$ is a trajectory of an object $i$ (in a given scene). The trajectory is defined by temporally ordered state vectors $s_{i,t}$, \ie, $\tau_i=(s_{i,t})_t$.  Each state $s_{i,t}$ encapsulates the attributes of the object %
(\eg, dimensions, heading) and center-point.%

We consider a motion forecasting model $f_\theta\colon s_{i,t-L:t} \mapsto s_{i,t+1:t+M}$, mapping $L$ past states and the current state of an object to its $M$ future states, parameterized by model weights $\theta$. {Other contextual information in the scene, \ie, other agents’ past trajectories and the map, can also be included in the input. We omit it for simplicity.} The model is optimized over the labeled trajectory set $\mathcal{T}$ as follows:
\begin{equation}
\textstyle
    \theta^* = \argmin_\theta\mathcal{L}(f_\theta; \mathcal{T}), \label{eq:optim-train}
\end{equation}
where $\mathcal{L}(f_\theta; \mathcal{T}) = \mathbb{E}_{\tau_i\sim p(\mathcal{T})}\mathbb{E}_{t\sim \mathcal{U}\{L,\operatorname{len}(\tau_i) - M\}}\ell(f_\theta(s_{i,t-L:t}),\allowbreak s_{i,t+1:t+M})$ is the prediction cost computed over the trajectory set. A commonly used loss $\ell$ is the Average Distance Error (ADE):
\begin{equation}
\textstyle
    \ell(f_\theta(s_{i,t-L:t}), s_{i,t+1:t+M}) = \frac{1}{M}\sum_{k=1}^M \| f(s_{i,t-L:t})_m - s_{i,t+m}\|_2^2.
\end{equation}
We notice that optimizing \autoref{eq:optim-train} relies solely on the labeled trajectory set $\mathcal{T}$, which implies a dependence on its labeling specificity.

\subsection{\method{} Training Strategy via Pseudo-Labeling}
\label{sec:create_ppt_dataset}
To reduce dependence on human-annotated trajectories, we propose a novel approach---\method{}---to leverage unannotated perception data and optionally labeled trajectories. \method{} uses perception data by generating raw, pseudo-labeled trajectories for pretraining motion forecasting models. We detail the PPT strategy in the following and illustrate it in \autoref{fig:ppt}.

\noindent\textbf{Pseudo-labeling via non-learning tracking.\;} We assume access to a perception dataset $\mathcal{D} = \{(o_{t})_t\}$, consisting of temporally ordered sensor data in different scenes at each time step $t$. Our goal is to construct a pseudo-labeled trajectory set $\tilde{\mathcal{T}}$ directly from this perception dataset $\mathcal{D}$. To achieve this,~\method{} leverages 3D multi-object tracking (MOT) methods to generate the pseudo labels from the sensor data sequence $(o_t)_t$. MOT methods typically follow the Tracking-by-Detection paradigm: At a given time $t$, a detector $\mathit{Dtr}\colon o_t\mapsto \delta_t$ processes the sensor data $o_t$ to produce a set of \textit{detected objects} $\delta_t = \{\tilde{s}_{k,t}\}_k$, each defined as the state vector mentioned above. We then initialize each track $i$ by assigning it an initial state $s_{i,0}\in\delta_0$. The tracker iteratively extends the trajectory by searching for the most likely object in the detected-object set $\delta_t$ at the next time step, \ie, $\mathit{Trk}\colon (s_{i,t-1}, \delta_t) \mapsto s_{i,t}$. 
{Current 3D object-tracking methods---like AB3DMOT \cite{Weng2020_AB3DMOT} and others---typically rely on a non-learned tracker for $\mathit{Trk}$, that links past tracks to detections via a one-to-one assignment. Consequently, these methods fundamentally operate as a form of pseudo-labeling for agent trajectories, since they have not been trained to \emph{predict} temporal waypoints.} 
After processing each scene in the perception dataset $\mathcal{D}$, we obtain a set of pseudo-labeled trajectories $\tilde{\mathcal{T}}$. %

\noindent\textbf{Pretraining with pseudo-labeled trajectories.\;}
Using the pseudo-labeling method described above, we can generate a variety of pseudo-labeled trajectory sets for any dataset, given a detector and tracker. Although these trajectories may not exactly match the human-annotated ones, they still serve well for pretraining. We provide a detailed analysis of their quality in \autoref{sec:pseudo_labels}, covering the datasets and detectors used in our experiments. Moreover, thanks to the abundance of available perception datasets, detectors, and trackers, it is convenient to scale this method to produce diverse trajectories with different domains, noise levels, and other variations beyond the labeled ground-truth annotations. To perform pretraining, we simply optimize \autoref{eq:optim-train} over our pseudo-labeled trajectory set \(\tilde{\mathcal{T}}\). That gives us pretrained weights \(\tilde{\theta}\) that capture the distribution of those pseudo-labeled trajectories. 

\noindent\textbf{Finetuning on human-annotated data.\;}
From there, one can optionally use the pretrained model as an initialization when training on the actual annotated data, such that it adapts to the specificity of the target domain for predictions. The finetuning still follows the same optimization in \autoref{eq:optim-train} but on a labeled trajectory subset in a chosen domain $\mathcal{T}'\subseteq \mathcal{T}$. This subset can be smaller than the original labeled set. Note that the finetuning is conducted independently of the data domains encountered during pretraining, enhancing the flexibility of the PPT strategy.

\section{Experimental Results}
\label{sec:exp}

We evaluate \method{} across a range of settings. \autoref{subsec:exp_setting} introduces the experimental setup. We compare \method{} with a `No-Pretraining' baseline under different annotation budgets in \autoref{subsec:ann_eff}, and assess its generalization in cross-domain, end-to-end, and multi-class settings in \autoref{subsec:generalization} and, \autoref{subsec:scalability} examines scalability. Further analyses are in \autoref{subsec:ablation} and some qualitative results are provided in \autoref{fig:qualitative}.

\subsection{Experimental Setting}\label{subsec:exp_setting}
\noindent\textbf{Datasets.\;} We use three {standard} \emph{perception} datasets\footnote{WOMD~\cite{womd} and AV2 MF~\cite{av2mf} datasets are omitted in our experiments due to the lack of off-the-shelf detectors and full perception data for the pseudo label extraction. We however show results on such datasets.}
with both raw sensors (for pseudo-labeling) and trajectory data (for baselines and finetuning) available. Namely, nuScenes (NUS)~\cite{nuscenes}, Waymo Open Dataset (WOD)~\cite{waymo} and Argoverse 2 (AV2)~\cite{wilson2023argoverse}.

\begin{table*}[t]
\scriptsize
\setlength{\tabcolsep}{2pt}

\centering
\caption{\textbf{\method{} \vs No Pretraining.} {Pretraining with \method{} consistently improves performance, especially with limited labeled data. 
Models are either trained from scratch (``No Pretraining'') or initialized with \colorbox{buscasota}{\method{}}. All models are then trained on 1\%, 10\%, or 100\% of labeled data.
All evaluations are on the full labeled valset.
Relative improvements (\%, \textcolor{teal}{teal}) are shown over ``No Pretraining''.
}
}

\begin{NiceTabular}{lc|cccc|cccc|cccc}
\toprule

\small\%  && \multicolumn{4}{c}{\textit{WOD} \cite{waymo}}  &  \multicolumn{4}{c}{\textit{AV2} \cite{wilson2023argoverse}}   & \multicolumn{4}{c}{\textit{NUS} \cite{nuscenes}} \\

labeled & \textbf{MTR} \cite{mtr}   
&Brier- & min- & min- & Miss-
&Brier- & min- & min- & Miss-
&Brier- & min- & min- & Miss-\\
trajs. &    
&FDE$\downarrow$ & ADE$\downarrow$ & FDE$\downarrow$ & Rate$\downarrow$ 
&FDE$\downarrow$ & ADE$\downarrow$ & FDE$\downarrow$ & Rate$\downarrow$ 
&FDE$\downarrow$ & ADE$\downarrow$ & FDE$\downarrow$ & Rate$\downarrow$ \\
\cmidrule(lr){1-2} \cmidrule(lr){3-6} \cmidrule(lr){7-10} \cmidrule(lr){11-14}

\multirow{3}{*}{\small 1\%}
&No Pretraining &     5.358 & 1.905 & 5.154 & 0.152

&   6.258 & 1.560 &5.768 & 0.214

&    6.039 & 2.411 & 5.764 & 0.209
 \\

&\rowcolor{buscasota}
\makecell{\method{} {(Ours)}\\ } & 
\makecell{{\best{0.585}}} &
\makecell{{\best{0.217}}} & 
\makecell{{\best{0.438}}} & 
\makecell{{\best{0.054}}} &

\makecell{{\best{1.181}}} & 
\makecell{{\best{0.399}}} & 
\makecell{{\best{0.913}}} & 
\makecell{{\best{0.098}}} &

\makecell{{\best{0.914}}} & 
\makecell{{\best{0.339}}} & 
\makecell{{\best{0.707}}} & 
\makecell{{\best{0.084}}}\\

 & &
\textcolor{teal}{-89\%}	& \textcolor{teal}{-89\%} &\textcolor{teal}{-92\%}	&\textcolor{teal}{-64\%} &
\textcolor{teal}{-81\%}	&\textcolor{teal}{-74\%}	&\textcolor{teal}{-84\%}	&\textcolor{teal}{-54\%} &
\textcolor{teal}{-85\%}	&\textcolor{teal}{-86\%}	&\textcolor{teal}{-88\%}	&\textcolor{teal}{-60\%} \\

\cmidrule(lr){1-2} \cmidrule(lr){3-6} \cmidrule(lr){7-10} \cmidrule(lr){11-14}

\multirow{3}{*}{\small 10\%}
&No Pretraining &  0.800 & 0.279 & 0.644 & 0.071&
 1.084 & 0.387 & 0.834 &0.097&

 1.310 & 0.472 & 1.109 &0.119\\

&\rowcolor{buscasota}
\makecell{\method{} {(Ours)}\\ } & 
\makecell{{\best{0.563}}} &
\makecell{{\best{0.209}}} & 
\makecell{{\best{0.419}}} & 
\makecell{{\best{0.050}}} &

\makecell{{\best{1.044}}} & 
\makecell{{\best{0.334}}} & 
\makecell{{\best{0.756}}} & 
\makecell{{\best{0.084}}} &

\makecell{{\best{0.856}}} & 
\makecell{{\best{0.325}}} & 
\makecell{{\best{0.650}}} & 
\makecell{{\best{0.082}}}\\

 & &
\textcolor{teal}{-30\%}	 &\textcolor{teal}{-25\%}	&\textcolor{teal}{-35\%}	&\textcolor{teal}{-30\%} & 
\textcolor{teal}{-4\%}	     &\textcolor{teal}{-14\%}	&\textcolor{teal}{-9\%}	&\textcolor{teal}{-13\%} &
\textcolor{teal}{-35\%}	&\textcolor{teal}{-31\%}	&\textcolor{teal}{-41\%}	&\textcolor{teal}{-31\%}\\

\cmidrule(lr){1-2} \cmidrule(lr){3-6} \cmidrule(lr){7-10} \cmidrule(lr){11-14}

\multirow{3}{*}{\small 100\%}
&No Pretraining &  0.562 & 0.219 & 0.416 & 0.052
&   0.905 &  0.307 & 0.644 &  0.072
&   0.923 &  0.327 & 0.722  &0.090\\

&\rowcolor{buscasota}
\makecell{\method{} {(Ours)}\\ } & 
\makecell{{\best{0.546}}} & 
\makecell{{\best{0.205}}} & 
\makecell{{\best{0.401}}} & 
\makecell{{\best{0.050}}} &

\makecell{{\best{0.894}}} & 
\makecell{{\best{0.291}}} & 
\makecell{{\best{0.624}}} & 
\makecell{{\best{0.069}}} &

\makecell{{\best{0.776}}} & 
\makecell{{\best{0.298}}} & 
\makecell{{\best{0.583}}} & 
\makecell{{\best{0.073}}}\\

& &
\textcolor{teal}{-3\%}	& \textcolor{teal}{-6\%}	& \textcolor{teal}{-4\%}	& \textcolor{teal}{-4\%} &
\textcolor{teal}{-1\%}	& \textcolor{teal}{-5\%}	& \textcolor{teal}{-3\%}	& \textcolor{teal}{-4\%} &
\textcolor{teal}{-16\%} & \textcolor{teal}{-9\%}	& \textcolor{teal}{-19\%}	&\textcolor{teal}{-19\%} \\

\bottomrule
\end{NiceTabular}
\vspace{-4mm}

\label{tab:pretrain}
\end{table*}
\noindent\textbf{Pseudo-labeled trajectories.\;} \xyh{To have pseudo-labeled trajectories from diverse sensor modalities (camera, LiDAR, multi-modal),} we collect detections from 9 state-of-the-art detectors and their variants~\cite{LiuTAYMRH23, YinZK21, BaiHZHCFT22, LiuTAYMRH23, YinSC0YFW24, wang2023dsvt, shi2023pv, chen2023voxelnext} and employ non-learning tracking methods~\cite{YinZK21, Weng2020_AB3DMOT} for detection association, which form 8.6M pseudo-labeled trajectories for \method{} pretraining. Without further specification, we focus on the 4-wheel vehicle objects. \xyh{Notably, we do not smooth or manually select the trajectories; all are results of the automated pipeline. This creates diverse feasible trajectories for each agent in the scene as demonstrated lately in~\autoref{fig:diverse_dataset}.}

\noindent\textbf{Forecasting method.\;}
To conduct experiments on different datasets, we leverage UniTraj~\cite{feng2024unitraj}, which provides a reliable implementation of multiple models, such as MTR~\cite{mtr}, Wayformer~\cite{Wayformer}, Autobot~\cite{autobot}, that predict 6-second future trajectories from 2-second past ground-truth history at 10 Hz.  Due to the page limit, our main results use MTR and additional results for Wayformer and Autobot are provided in \autoref{sec:add_res}. %

\xyh{\noindent\textbf{Training.\;\;}We pretrain the forecasting models for 16 epochs (in practice, convergence may be obtained before). For finetuning, we train for 40 epochs and report early-stopping scores based on validation evaluation. We follow the hyper-parameter setting of UniTraj~\cite{feng2024unitraj} in all experiments.
For finetuning, the learning rate is reduced by an order of magnitude of 0.1; the other hyperparameters remain unchanged.}

\noindent\textbf{Evaluation metrics.\;}
We evaluate the predicted trajectories with respect to the ground-truth future trajectories with: \begin{itemize*} \item $\text{minADE}_k$: The minimum Average Displacement Error over $k$ predictions, computed as the mean $L^2$ distance between each predicted trajectory location and the ground-truth one. 
\item $\text{minFDE}_k$: The minimum Final Displacement Error over $k$ predictions, which measures the $L^2$ distance between the predicted endpoint and the ground truth. 
\item $\text{Brier-FDE}$: The sum of $\text{minFDE}_k$ and $(1 - \delta)^2$, where $\delta$ represents the forecasting confidence score, quantifying the model's confidence in the trajectory prediction that most closely matches the ground truth.
\item $\text{MissRate}_{k@x}$: The proportion of forecasts where $\text{minFDE}_k$ exceeds a threshold $x$, set to $2$ meters. 
\end{itemize*} We set $k=6$ for all metrics.

\begin{table}[h]
\centering
\caption{\textbf{Quality Assessment of Pseudo-Labeled Trajectories.} We show here the MF metrics as the proxy to demonstrate the quality of our pseudo-labeled trajectories used for pretraining.
}
\label{tab:tracking_perf}
\small
\setlength{\tabcolsep}{5pt}
\resizebox{0.8\linewidth}{!}
{
\begin{tabular}{@{}clccc@{\hspace{0.3cm}}ccc@{}}
\toprule
&& \multicolumn{3}{c}{\text{Trainset}} & \multicolumn{3}{c}{\text{Valset}} \\
\midrule
&{ \textbf{Tracker}} &  \text{min-}  &  \text{min-} &  \text{Miss-}  &  \text{min-}  &  \text{min-} &  \text{Miss-} \\
&&  \text{ADE$\downarrow$}  &  \text{FDE$\downarrow$} &  \text{Rate$\downarrow$ }  &  \text{ADE$\downarrow$}  &  \text{FDE$\downarrow$} &  \text{Rate$\downarrow$ } \\

\cmidrule(l){2-8}
\multirow{-2}{*}{\rotatebox[origin=c]{90}{\small\textbf{ AV2}}} & BEVFusion ~\cite{LiuTAYMRH23}&  0.274 & 0.375& 0.030 & 0.295 & 0.404 & 0.032\\

\midrule
\multirow{4}{*}{\rotatebox{90}{\textbf{NUS}}}
&CenterPoint~\cite{YinZK21}& 0.402&0.593 & 0.090 &  0.414&0.600  &0.080\\
&TransFusion-L~\cite{BaiHZHCFT22}& 0.266 & 0.326& 0.040  & 0.264 & 0.323& 0.034\\
&BEVFusion~\cite{LiuTAYMRH23}& 0.256 & 0.310 & 0.035 & 0.250 & 0.298 &  0.026\\
&IS-Fusion~\cite{YinSC0YFW24}& 0.256 & 0.312& 0.037 &0.252 & 0.305 &  0.027\\

\midrule
\multirow{4}{*}{\rotatebox{90}{\textbf{WOD}}}
&DSVT-Pillar~\cite{wang2023dsvt}&  0.383 & 0.366&  0.015 & 0.385& 0.365& 0.014 \\
&DSVT-Voxel~\cite{wang2023dsvt}& 0.378 & 0.362 & 0.014 & 0.379& 0.360 & 0.014 \\
&PV-RCNN++~\cite{shi2023pv}& 0.237  & 0.247 &  0.011 & 0.245& 0.252 & 0.011\\
&VoxelNeXt~\cite{chen2023voxelnext}& 0.564 & 0.532 &  0.042& 0.548& 0.511 & 0.039\\

\midrule

&\textbf{Average}& \textbf{0.297} & \textbf{0.380} &  \textbf{0.035} & \textbf{0.337}& \textbf{0.379} & \textbf{0.031}\\
\bottomrule
\end{tabular}
}
\vspace{-4mm}
\end{table}

\subsection{Pseudo Labels Quality Analysis}\label{sec:pseudo_labels}
\xyh{In \autoref{tab:tracking_perf}, we assess the forecasting performance of the collected tracking trajectories. Precisely, at each time step of a given sequence, we perform a matching between predicted and ground truth based on their past trajectories with a threshold of 2 meters. After that, we calculate forecasting metrics. In general, the tracking trajectories deviate from the ground truth annotations by less than 0.5 meters on average, generally lower than the average forecast performance of the model used and trained with labeled data. Although they do not fully match the quality of labeled annotations, we show in the following experiments that inexpensive and noisy trajectories are beneficial for motion forecasting pretraining, \eg, when ground truth is absent or limited.}

\subsection{Annotation Efficiency} \label{subsec:ann_eff}
In the section, we show that~\method{} generally improves the baselines via the pretraining strategy with diverse off-the-shelf trajectories. This performance gap is especially pronounced {when only a small fraction of labeled data is available, demonstrating the strong annotation efficiency of \method{}}.

\label{sec:resultss}
\noindent\textbf{\method{} \vs No Pretraining in different labeled data regimes.\;}
\autoref{tab:pretrain} reports the results of single-dataset training where the training and validation are done on the same dataset. {Models trained from scratch (`No Pretraining') consistently underperform compared to those initialized with \method{}, especially in low-label regimes (1\%, 10\%). Notably, models pretrained with \method{} and finetuned on just 10\% of labeled data match or surpass models trained from scratch on 100\% of labeled data. In the extreme case of using only 1\% of labeled trajectories, the performance gap becomes even larger.}

Indeed, the pretrained models ingest extra pseudo-labeled trajectories; but those labels are generated automatically (at scale and at negligible cost) from \textit{the same sensor data} used for human annotations, ensuring a \textit{fair comparison}. Importantly, this inexpensive pretraining dataset significantly reduces reliance on costly, manual annotations, demonstrating that large quantities of automatically generated trajectories can meaningfully boost forecasting performance.

\noindent\textbf{\method{}~\textit{w/o} finetuning for adaptation.\;}{We also evaluate models trained \emph{only} with pseudo-labeled data, without any supervised finetuning.
As shown in \autoref{tab:id_noisy_ood_clean}, pretraining on in-domain pseudo-labeled trajectories outperforms models trained on human-labeled but out-of-domain datasets.
This result is particularly useful for practitioners working with new domains without labeled data.
In such cases, \method{} offers a compelling solution: train directly on pseudo labels from the target domain instead of transferring from unrelated labeled data, which often suffers from domain gaps.
}

\begin{table}[t]
\setlength{\tabcolsep}{3pt}
\begin{minipage}[t]{0.49\linewidth}
\centering
\caption{\textbf{PPT \emph{w/o} finetuning for adaptation.} {Pretraining only on in-domain pseudo labels outperforms training on human-labeled data from a different domain.}}
\label{tab:id_noisy_ood_clean}
\small
\resizebox{\linewidth}{!}{%
\begin{NiceTabular}{clccccc}
\toprule

\multirow{2}{*}{\rotatebox{90}{Target}}
& & & Brier- & min- & min- & Miss-   \\
& & Source & FDE$\downarrow$ & ADE$\downarrow$ & FDE$\downarrow$ & Rate$\downarrow$   \\
\cmidrule{2-7}

 &\multirow{2}{*}{No Pretraining} &   AV2 &   0.886	& 0.318	& 0.718	& 0.063\\
 & &   NUS  &  1.539 & 0.577&	1.362&	0.093 \\
 \rowcolor{buscasota} \cellcolor{white}
\multirow{-3}{*}{\rotatebox{90}{\textbf{WOD}}}&
\makecell{\method{} (Ours)\\ } & pseudo WOD & 
\makecell{{\best{0.586}}} & 
\makecell{{\best{0.229}}} & 
\makecell{{\best{0.449}}} & 
\makecell{{\best{0.055}}} \\

\midrule

  &\multirow{2}{*}{No Pretraining} &  WOD  & {\best{1.302}}	& 0.441	& 1.007	&0.118  \\
  & &  NUS &  1.483 & 0.529& 1.205 &	0.124 \\
\rowcolor{buscasota} \cellcolor{white}
\multirow{-3}{*}{\rotatebox{90}{\textbf{AV2}}}&
\makecell{\method{} (Ours)\\ } & pseudo AV2 & 
\makecell{1.311} & 
\makecell{{\best{0.430}}} & 
\makecell{{\best{0.995}}} & 
\makecell{{\best{0.102}}} \\

\midrule

 &\multirow{2}{*}{No Pretraining} &  WOD   &  1.427	& 0.625&	1.152	& 0.150  \\
 &&  AV2  &  1.128	&0.557	&0.882	& 0.105 \\
 \rowcolor{buscasota} \cellcolor{white}
 \multirow{-3}{*}{\rotatebox{90}{\textbf{NUS}}}&
\makecell{\method{} (Ours)\\ } &  pseudo NUS& 
\makecell{{\best{1.003}}} & 
\makecell{{\best{0.356}}} & 
\makecell{{\best{0.793}}} & 
\makecell{{\best{0.089}}} \\

\bottomrule
\end{NiceTabular}
}

\end{minipage}\hfill
\begin{minipage}[t]{0.49\linewidth}
\caption{\textbf{Cross-domain generalization.} {\colorbox{buscasota}{\method{}} pretraining with diverse pseudo-labeled trajectories improves cross-domain generalization.}
}
\label{tab:cross_dataset_evaluation}
\resizebox{\linewidth}{!}{%
    \begin{tabular}{@{}c@{\hskip 4pt}lcc|cccc@{}}
\toprule

\multirow{2}{*}{\rotatebox{90}{Target}}
&& Pre- &Fine-&Brier- & min- & min- & Miss-   \\
&& train & tune &FDE$\downarrow$ & ADE$\downarrow$ & FDE$\downarrow$ & Rate$\downarrow$   \\
\cmidrule{2-8}

&No Pretraining & $\times$ &   &  0.886	& 0.318	& 0.718	& 0.063\\
\rowcolor{buscasota} \cellcolor{white}&
\method{} (Ours) & pseudo AV2 & \multirow{-2}{*}{AV2} &
\best{0.798} & 
\best{0.310} & 
\best{0.611} & 
\best{0.053} \\

\cmidrule{2-8}

&No Pretraining &$\times$&  &  1.539 & 0.577&	1.362&	0.093\\
\rowcolor{buscasota}
\cellcolor{white}\multirow{-4}{*}{\rotatebox{90}{\textbf{WOD}}}&
\method{} (Ours) & pseudo NUS  & \multirow{-2}{*}{NUS} &
\best{0.988} &
\best{0.371} & 
\best{0.836} & 
\best{0.079} \\

\midrule

&No Pretraining & $\times$&   & {\best{1.302}}	& {\best{0.441}}	& {\best{1.007}}	&0.118 \\
\rowcolor{buscasota} \cellcolor{white}&
\method{} (Ours) & pseudo WOD & \multirow{-2}{*}{WOD} &
1.363 & 
0.445 & 
1.074 & 
\best{0.109} \\

\cmidrule{2-8}

&No Pretraining &$\times$ &  &  1.483 & 0.529&1.205&	0.124 \\
\rowcolor{buscasota}\cellcolor{white}
\multirow{-4}{*}{\rotatebox{90}{\textbf{AV2\!\!}}}&
\method{} (Ours) & pseudo NUS & \multirow{-2}{*}{NUS} &
\best{1.453} &
\best{0.485} & 
\best{1.182} & 
\best{0.114} \\

\midrule

&No Pretraining & $\times$&   &  1.427	& 0.625&	1.152	& 0.150 \\
\rowcolor{buscasota} \cellcolor{white}&
\method{} (Ours) & pseudo WOD & \multirow{-2}{*}{WOD} &
\best{1.207} & 
\best{0.458} & 
\best{0.959} & 
\best{0.102} \\

\cmidrule{2-8}

&No Pretraining & $\times$ &  &  1.128	&0.557	&0.882	& 0.105 \\
\rowcolor{buscasota} \cellcolor{white}
\multirow{-4}{*}{\rotatebox{90}{\textbf{NUS}}}&
\method{} (Ours)& pseudo AV2 & \multirow{-2}{*}{AV2} &
\best{1.071} & 
\best{0.470} & 
\best{0.822} & 
\best{0.093} \\

\bottomrule
\end{tabular}
}
\vspace{-4mm}

\end{minipage}
\end{table}

\begin{figure}[h]
\centering
\subfloat[WOD \cite{waymo}]{
\includegraphics[trim={1.2cm 0cm 0 0cm},clip, width=0.32\linewidth]{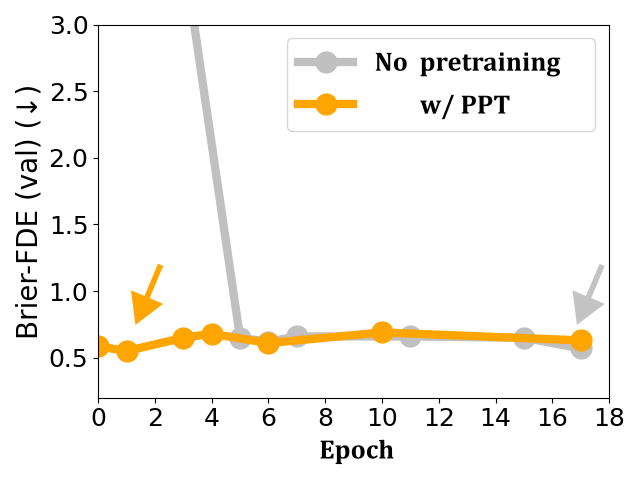}}
\hspace*{\fill}
\subfloat[Argoverse 2 \cite{wilson2023argoverse}]{
\includegraphics[trim={1.2cm 0cm 0 0cm},clip, width=0.32\linewidth]{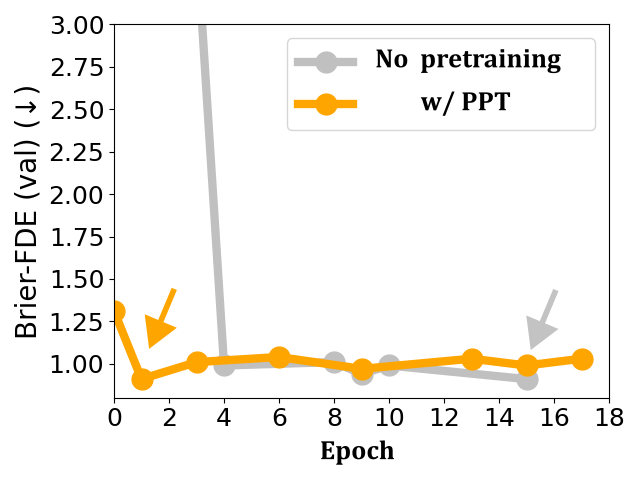}}
\hspace*{\fill}
\subfloat[nuScenes \cite{nuscenes}]{
\includegraphics[trim={1.3cm 0cm 0 0cm},clip, width=0.32\linewidth]{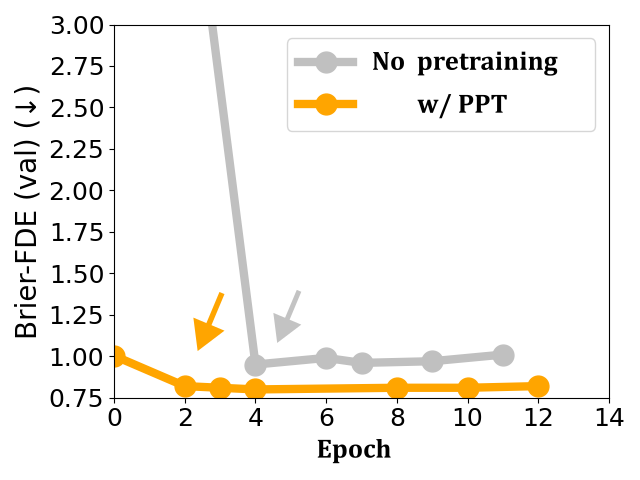}}
\caption{\textbf{Efficient finetuning}. We show the evolution of $\text{brier-FDE}$ on the validation sets during \textcolor{gray}{training from scratch}/\textcolor{orange}{finetuning w/~\method{}} the forecasting MTR model. Models pretrained with PPT converges faster and achieves better results (indicated by arrows).}
\label{fig:ft_cost}
\end{figure}

\noindent\textbf{Efficient supervised finetuning.}
\xyh{Furthermore, we show in \autoref{fig:ft_cost} that finetuning after~\method{} pretraining converges much faster, typically in only 1 or 2 epochs. For reference, one epoch takes around 16 minutes on WOD~\cite{waymo} with the MTR model. These results suggest that most of the forecasting ability was acquired during the pretraining phase, with minimal adaptation required to adjust the model to the labeled data of the benchmarks.}

\subsection{Improved Generalization with \method{}}\label{subsec:generalization}

\noindent\textbf{Cross-domain generalization with \method{}.} {Pretraining on diverse (\autoref{fig:diverse_dataset}) pseudo-labeled trajectories from a single dataset leads to better generalization to unseen domains. As shown in \autoref{tab:cross_dataset_evaluation}, models pretrained with \method{} on pseudo-labeled data from a source dataset outperform those trained from scratch when evaluated on a different target dataset.
For instance, when we pretrain a model using pseudo-labeled trajectories from the NUS dataset (via \method{}), then finetune it on the human-annotated labels of NUS, and finally evaluate it on the WOD dataset, it outperforms a model that was trained only on the labeled NUS data. We attribute this to the noise and diversity of pseudo-labeled trajectories, which act as regularization and expose the model to a broader distribution of behaviors, ultimately improving robustness and generalization.
}

\begin{table}[h]
\setlength{\tabcolsep}{3pt}
\centering
\caption{\textbf{E2E motion forecasting on AV2.} With \colorbox{buscasota}{\method{}}, the end-to-end forecasting model~\cite{xu2024valeo4cast} becomes more robust against imperfect perception inputs in the end-to-end forecasting paradigm. \textcolor{teal}{Teal} shows relative gains (\%) over `No Pretraining'.
}
\label{tab:e2eforcasting}
\resizebox{0.9\linewidth}{!}{%
\begin{NiceTabular}{lc|ccc}
\toprule
\textbf{Modular4Cast}~\cite{xu2024valeo4cast}&  &\multicolumn{3}{c}{AV2 E2E MF} \\
 & Source& $\text{mAP}_{f}\uparrow$ & minADE$\downarrow$ & minFDE$\downarrow$\\
\midrule

 No Pre-training & AV2 &0.367	& 2.868	& 4.807  \\

 \rowcolor{buscasota}
\method{} (Ours)& pseudo AV2 & \textbf{0.595}	& \textbf{0.874}	&  \textbf{1.430} \\

& & \textcolor{teal}{+62.13\%}	& 	\textcolor{teal}{-69.53\%}& \textcolor{teal}{-70.25\%} \\

\bottomrule
\end{NiceTabular}
}
\end{table}
\begin{table}[h]
\scriptsize
\setlength{\tabcolsep}{2pt}
\vspace{-2mm}
\centering
\caption{\textbf{Multi-class motion forecasting.}  \colorbox{buscasota}{\method{}} boosts performance on the AV2 MF~\cite{av2mf} multi-class benchmark. Evaluated on the testset via the official server. \textcolor{teal}{Teal} shows relative gains (\%) over `No Pretraining'.
}\label{tab:av2mftest}

\resizebox{0.8\linewidth}{!}{%
\begin{NiceTabular}{cc|cccc}
\toprule

  & & \multicolumn{4}{c}{AV2 MF Testset} \\

&&Brier-FDE$\downarrow$  & minADE$\downarrow$ & minFDE$\downarrow$  & MissRate$\downarrow$\\
\midrule
&No Pretraining &   2.282 & 0.832 & 1.658 & 0.286\\
&\rowcolor{buscasota} \makecell{\method{} {(Ours)}\\ } & \makecell{{\best{2.246}}} & \makecell{{\best{0.820}}} & 
\makecell{{\best{1.630}}} & 
\makecell{{\best{0.272}}}\\

  & &
\textcolor{teal}{-1.58\%}	&\textcolor{teal}{-1.44\%}	&\textcolor{teal}{-1.69\%}	&\textcolor{teal}{-4.90\%} \\

\bottomrule
\end{NiceTabular}
}
\vspace{-4mm}

\end{table}

\noindent\textbf{Toward end-to-end forecasting with \method{}.\;} {End-to-end (E2E) forecasting uses \textit{predicted} past trajectories—generated by a perception model—as input, reflecting real-world deployment where clean, labeled inputs are unavailable~\cite{peri2022forecasting_future_detection,xu2024towards,xu2024valeo4cast}.
To jointly evaluate perception and forecasting, the metric $\text{mAP}_{f}$~\cite{peri2022forecasting_future_detection} is used, similar to the detection AP, matching predicted future trajectories to ground truth.} We integrate \method{} into the winning solution~\cite{xu2024valeo4cast} of the AV2 E2E forecasting challenge, training the model either on labeled AV2 data (No pretraining) or solely on pseudo-labeled AV2 data~\cite{LiuTAYMRH23} (\method{} without finetuning).
{
As shown in \autoref{tab:e2eforcasting}, the model trained with \method{} outperforms its supervised counterpart, despite using no ground-truth labels. This highlights \method{} as a promising strategy for improving forecasting robustness in E2E settings with the presence of imperfect perception.
}

\noindent\textbf{Toward multi-class motion forecasting with \method{}.\;}
We extend our evaluation to a multi-class setting, predicting trajectories for 10 distinct agent types as defined in the AV2 Motion Forecasting (AV2 MF) benchmark~\cite{av2mf}: {\footnotesize \texttt{VEHICLE}, \texttt{PEDESTRIAN}, \texttt{MOTORCYCLIST}, \texttt{CYCLIST}, \texttt{BUS}, \texttt{BACKGROUND}, \texttt{STATIC},  \texttt{CONSTRUCTION}, \texttt{RIDERLESS\_BICYCLE} and \texttt{OTHER}}.
To do this, we pretrain the MTR model~\cite{mtr} with \method{} on pseudo-labeled data of the perception datasets~\cite{nuscenes, wilson2023argoverse,  waymo}, including all classes, and then finetune it on the AV2 MF~\cite{av2mf} training set.
We evaluate by submitting predictions to the AV2 MF test server. As shown in \autoref{tab:av2mftest}, pretraining with \method{} on pseudo-labeled perception datasets also improves performances in standard motion forecasting benchmarks in the multi-class setting.

\subsection{Scaling Pretraining and Finetuning} \label{subsec:scalability}

We study how performance changes when increasing the amount of pretraining or finetuning data. We note that we use the \textit{same} amount of data (i.e., the same number of agents) in pseudo-labeled trajectories and their ground-truth counterparts, but the pseudo-labeled trajectories are extracted from different detectors, automatically creating diverse and realistic training examples. {In the setting noted `Single', the pretraining or finetuning is done on the same labeled data as the evaluation. The model is then evaluated on the same dataset, and we report the average performance across the three datasets.}
In the setting noted `All', we combine the three motion forecasting datasets~\cite{nuscenes, wilson2023argoverse,  waymo} for the pretraining or finetuning. {The evaluation is then performed on the respective evaluation sets of these datasets, and the average performance is reported.}

\noindent\textbf{Pretraining on more pseudo-labeled data.\;} {A key advantage of \method{} is that pseudo-labeled trajectories generated from different datasets are naturally compatible: no manual annotation, alignment, or post-processing is required. This makes it straightforward to combine data from multiple sources, something that is far more challenging with labeled datasets due to differing formats and annotation protocols \cite{feng2024unitraj}.
As shown in \autoref{tab:multi-dataset}, pretraining on the combined dataset (`All') consistently outperforms pretraining on any single dataset. This scaling effect highlights the value of large and diverse pseudo-labeled data in improving motion forecasting performance, and further increases the gap between models trained from scratch and those using \method{}.
}

\begin{table}[h]
\setlength{\tabcolsep}{3pt}
\centering
\caption{\textbf{Scaling pretraining and finetuning.}
Comparison of models trained from scratch (`No Pretraining') versus those using \colorbox{buscasota}{\method{}} pretraining on a single dataset (`Single') or all three combined (`All'). {We report the average evaluation metrics computed separately on each dataset.} Relative gains (\%, \textcolor{teal}{in teal}) are given against the `No Pretraining' baseline.
} \label{tab:multi-dataset}
\resizebox{\linewidth}{!}{%
\begin{NiceTabular}{lcc|cccc}
\toprule
&&&  \multicolumn{4}{c}{\textit{Average} } \\
&Pretrain& Finetune &\smaller Brier-FDE$\downarrow$  & \smaller minADE$\downarrow$  & \smaller minFDE$\downarrow$  & \smaller MissRate$\downarrow$   \\

\cmidrule(lr){4-7}

No Pretraining & $\times$& Single & 0.797	&0.284	&0.594	& 0.071\\
\rowcolor{buscasota}
\makecell{\method{} (Ours)\\ } &Single& Single&
\makecell{{{0.739}}} & 
\makecell{{{0.265}}} & 
\makecell{{{0.536}}} & 
\makecell{{{0.064}}}\\
\rowcolor{buscasota}									
\makecell{\method{} (Ours)\\ } &All& Single&
\makecell{{\best{0.700}}} & 
\makecell{{\best{0.251}}} & 
\makecell{{\best{0.507}}} & 
\makecell{{\best{0.061}}} \\
& & & \textcolor{teal}{-12.18\%} & \textcolor{teal}{-11.61\%}	& \textcolor{teal}{-14.59\%}	&\textcolor{teal}{-14.49\%} \\

\midrule
No Pretraining & $\times$& All & 0.739	&0.295	&0.553	& \best{0.060} \\

\rowcolor{buscasota}
\makecell{\method{} (Ours)\\ } &All& All&
\makecell{{\best{0.713}}} & 
\makecell{{\best{0.264}}} & 
\makecell{{\best{0.522}}} & 
\makecell{{{0.062}}}\\
&&&\textcolor{teal}{-3.52\%} &\textcolor{teal}{-10.51\%} &\textcolor{teal}{-5.61\%} &\textcolor{teal}{+3.33\%} \\
\bottomrule
\end{NiceTabular}
}
\end{table}

\noindent\textbf{Finetuning on more labeled data.}
While \method{} consistently outperforms the baseline when both pretraining and finetuning use the combined datasets (`All'), increasing the amount of labeled data for finetuning does not always yield additional gains compared to using only a `Single' dataset for finetuning. We hypothesize that the model already captures strong motion priors from diverse pseudo-labeled trajectories during the pretraining phase. Finetuning then mainly serves as a lightweight adaptation step, helping the model align with human-labeled labels. This is supported by our analysis in~\autoref{subsec:ablation}, where we observe that performance often plateaus after only 1–2 epochs of finetuning.

\subsection{Further Analyses}
\label{subsec:ablation}

\noindent\textbf{Choice of 'good' and 'bad' trajectories.} In PPT, we deliberately select trackers from diverse modalities. To study how the quality of pseudo-labeled trajectories impacts the forecasting performance, \xyh{we evaluated on WOD (15k samples), as shown~\autoref{tab:good_bad_detectors}, where we remove one or two detectors from the training set. ‘Past ADE’ gives the quality of input past trajectories from the removed tracks. Increases in ‘Future ADE’ are w.r.t. \textit{PPT (all detectors)}. We observe that: (1) reducing the number of detectors causes a systematic performance drop; (2) removing the best-performing detectors has a bigger negative impact than the worse-performing ones.}
\begin{table}
\caption{We uniformly sample 15,290 pseudo-labeled trajectories from 100 WOD~\cite{waymo} scenarios of different detectors and MTR~\cite{mtr} models are pretrained without finetuning. Keeping the same amount of data, we remove one or two detectors from the training set ('DSVT-P' refers to DSVT-Pillar and 'DSVT-V' refers to DSVT-Voxel~\cite{wang2023dsvt}). All models are evaluated on the full validation set of WOD. ADE is reported since it reflects the overall quality of the trajectories.}\label{tab:good_bad_detectors}
\resizebox{\linewidth}{!}{%
\begin{tabular}{lcccccccc}
\toprule
 & w/o VoxelNext (a) & w/o DSVT-P (b) & w/o DSVT-V (c) & w/o PV-RCNN++ (d) \\
\midrule
\textbf{Past ADE↓ (Tab.9)} & 0.564 & 0.383 & 0.378 & 0.237  \\
\textbf{PPT -- Future ADE↓} & 0.305 (+0.032) & 0.319 (+0.046) & 0.311 (+0.038) & 0.350 (+0.077)  \\
 \toprule
 & w/o (b) and (c) & w/o (d) and (c) & w/o (a) and (b) & \cellcolor{buscasota} PPT (all detectors)\\
\midrule
\textbf{Past ADE↓ (Tab.9)} & 0.378/0.383 & 0.237/0.378 & 0.564/0.383 & \cellcolor{buscasota}- \\
\textbf{PPT -- Future ADE↓} & 0.323 (+0.05) & 0.376 (+0.103) & 0.322 (+0.049)& \cellcolor{buscasota} \bf 0.273 \\
\bottomrule
\end{tabular}
}
\end{table}

\begin{figure}[t]
\setlength{\tabcolsep}{3pt}
\begin{minipage}{0.45\linewidth}
\centering
\centering
\subfloat[Curated traj.]{
\includegraphics[trim={0.0cm 0cm 0 0cm},clip, width=0.45\linewidth]{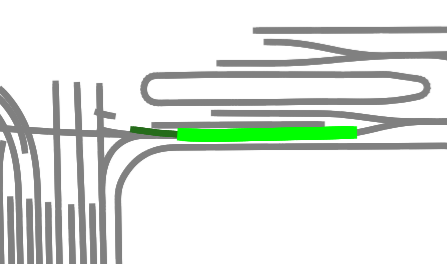}}
\subfloat[DSVT-Voxel]{
\includegraphics[trim={0.0cm 0cm 0 0cm},clip, width=0.45\linewidth]{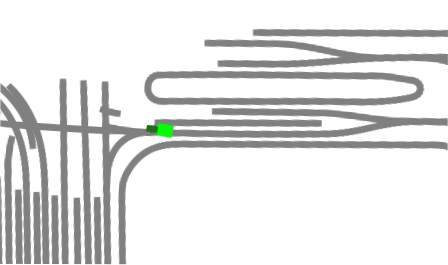}}

\subfloat[PV-RCNN++]{
\includegraphics[trim={0.0cm 0cm 0 0cm},clip, width=0.45\linewidth]{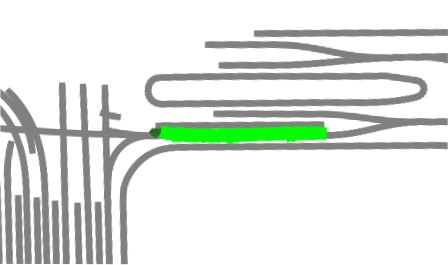}}
\subfloat[VoxelNeXt]{
\includegraphics[trim={0.0cm 0cm 0 0cm},clip, width=0.49\linewidth]{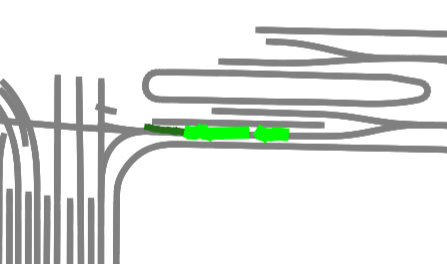}}

\caption{\textbf{Diverse pseudo-labeled trajectories.} Compared to a single curated annotation in WOD~\cite{waymo}, with pseudo-labeled trajectories, we provide not only the training example close to the ground truth but other feasible trajectories.}
\label{fig:diverse_dataset}
\end{minipage}\hfill
\begin{minipage}{0.49\linewidth}
\centering
\setlength{\tabcolsep}{3pt}
\subfloat{
\includegraphics[trim={0.0cm 0cm 0 0cm},clip, width=\linewidth]{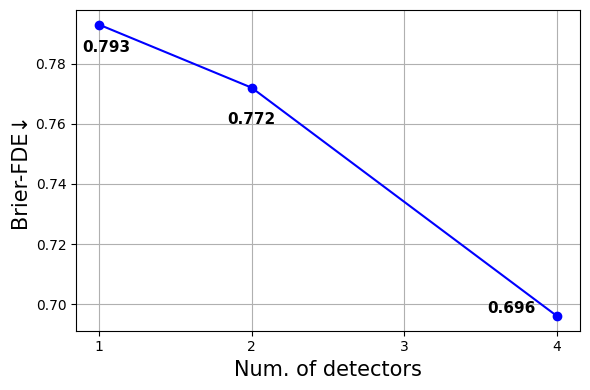}}
\caption{\textbf{Data diversity in pretraining matters.} MTR models are pretrained without finetuning with exactly 15,290 samples from 100 scenarios of WOD~\cite{waymo}. These samples come from one, two or four detection models. All models are evaluated on the full validation set of WOD.
}
\label{fig:data_diversity_amount}
\end{minipage}
\vspace{-4mm}
\end{figure}

\noindent\textbf{Variability of trajectories matters.}
{
\xyh{Besides the noise level of each pseudo-labeled trajectory, PPT does not select a single trajectory for each agent but maintains diverse feasible trajectories for pretraining, as shown in~\autoref{fig:diverse_dataset}, which we refer as trajectory variability.} We hypothesize that this variability acts as a strong form of regularization, helping models learn richer and more generalizable motion representations.
To quantify how the trajectory variability impacts forecasting performance, we conduct a controlled experiment where we fix the scenarios and the total number of pretraining trajectories and vary only the number of source trackers. Specifically, on the same 100 WOD scenarios, we uniformly sample 15,290 pseudo-labeled trajectories from (i) a single detector~\cite{shi2023pv}, (ii) two different detectors~\cite{chen2023voxelnext, shi2023pv}, or (iii) four detectors~\cite{chen2023voxelnext, shi2023pv, wang2023dsvt}\footnote{DSVT \cite{wang2023dsvt} includes two variants: DSVT-Voxel and DSVT-Pillar.}. We then pretrain MTR~\cite{mtr} using these different pseudo-labeled trajectories and evaluate the impact to motion forecasting. As shown in~\autoref{fig:data_diversity_amount}, models pretrained on more diverse pseudo-labeled trajectories consistently outperform those trained on data from fewer sources. These results show the crucial role of variability in pseudo-labeled trajectories for better effective motion forecasting pretraining.
}

\noindent\textbf{Post-processing is unnecessary for pre-training.} \xyh{To investigate the {role of post-processing} (`pp') in pseudo-labeling more rigorously, we compare pseudo-labels with and without `pp'~\cite{zhang2022bytetrack} on the same dataset. We show in~\autoref{tab:no_pp} that: (1) PPT vs.~PPT+pp, the post-processing is not necessary; (2) PPT vs.~LION~\cite{liu2024lion}+pp, a simplified emulation of the labeling process in WOMD~\cite{womd} does not exhibit higher performance.} 

\begin{table}
\caption{We pretrain two MTR models: (1) A state-of-the-art tracker Lion~\cite{liu2024lion}+`pp'; (2) PPT+`pp'.
The finetuning is conducted in either the same (i.e., in-domain, top) or different (i.e., cross-domain, bottom) datasets. The performance is averaged over the datasets~\cite{waymo, nuscenes, wilson2023argoverse}.}~\label{tab:no_pp}
\centering
\resizebox{0.8\linewidth}{!}{%
\begin{tabular}{@{}l|cccc@{}}
\toprule
 & \textbf{B-FDE}↓ & \textbf{minADE}↓ & \textbf{minFDE}↓ & \textbf{MissRate}↓ \\
 \toprule
&  \multicolumn{4}{c}{\textit{Average (in-domain)}} \\
\midrule
No pretraining  & 0.797 & 0.284 & 0.594 & 0.071 \\

\rowcolor{white}
LION + pp & 0.779 &	0.283	&0.582	&0.069   \\

\rowcolor{buscasota}
PPT   & \textbf{0.739} & \textbf{0.265} & \textbf{0.536} & \textbf{0.064} \\

PPT + pp  & 0.773 & 0.278 & 0.571 & 0.067 \\
\toprule
& \multicolumn{4}{c}{\textit{Average (cross-domain)}} \\
\midrule
No pretraining   &  1.294	&0.508	& 1.054 & 0.114\\

\rowcolor{white}
LION + pp &  1.192 &	0.455&	0.966 & 0.095\\

\rowcolor{buscasota}
PPT  &  \textbf{1.147}	&\textbf{0.438}	&\textbf{0.924} & \textbf{0.096}\\

PPT + pp & 1.183	& 0.447	& 0.953	& 0.094 \\
\bottomrule
\end{tabular}
}
\vspace{-4mm}
\end{table}

\begin{table}[h]
\caption{\textbf{Pretraining \textit{w/o} HD maps.} Using~\colorbox{buscasota}{\method{}}, we pretrain the forecasting model \textit{w/} or \textit{w/o} HD maps, compared to no pretraining. We note that all experiments pretrain and finetune using `All' three datasets and HD maps are used during the finetuning phase. Relative gains (\%, \textcolor{teal}{in teal}) are given against the `No Pretraining' baseline. 
}\label{tab:multi-dataset_result-map}
\setlength\extrarowheight{-3pt}
\centering
\resizebox{\linewidth}{!}{%
\begin{NiceTabular}{lcc|cccc}
\toprule

&&&  \multicolumn{4}{c}{\textit{Average} } \\
&Pretrain& Finetune &\smaller Brier-FDE$\downarrow$  & \smaller minADE$\downarrow$  & \smaller minFDE$\downarrow$  & \smaller MissRate$\downarrow$   \\

\cmidrule(lr){4-7}
No Pretraining & $\times$& All & 0.739	&0.295	&0.553	& {0.060} \\

\rowcolor{buscasota}
\makecell{\method{} (Ours)\\ } &All& All& \makecell{{{0.713}}} &  \makecell{{\best{0.264}}} &  \makecell{{\best{0.522}}} &  \makecell{{{0.062}}}\\
&&&\textcolor{teal}{-3.52\%} &\textcolor{teal}{-10.51\%} &\textcolor{teal}{-5.61\%} &\textcolor{teal}{+3.33\%} \\
\rowcolor{buscasota}	
\method{} $w/o$ Maps (Ours) & All & All &  \best{0.706}	&0.266&	0.528&	\best{0.059}\\
&& &\textcolor{teal}{-4.50\%}	& \textcolor{teal}{-10.06\%}	&\textcolor{teal}{-4.62\%}	& \textcolor{teal}{-2.19\%}  \\
\bottomrule
\end{NiceTabular}

}
\end{table}
 
\noindent\textbf{Pretraining \textit{w/o} HD maps.\;} 
{
In all previous experiments, we assume the availability of HD maps, which are commonly used in motion forecasting research, to isolate the effect of pseudo-labeled trajectories during pretraining. In \autoref{tab:multi-dataset_result-map}, we relax this assumption by pretraining the model without HD maps, using only the pseudo-labeled trajectories. Specifically, we use empty HD maps during pretraining. The results show that pretraining without HD maps achieves performance on par with the pretraining that incorporates them. This suggests that the primary benefit of pretraining comes from the trajectory dynamics and agent interactions, rather than the agent-map interactions.
}

\begin{figure}[h] 
\centering
\setlength{\tabcolsep}{2pt}
\resizebox{0.8\linewidth}{!}{%
\begin{tabular}{c}
\includegraphics[width=0.8\linewidth,trim={0.0cm 1.5cm 0cm 0cm},clip]{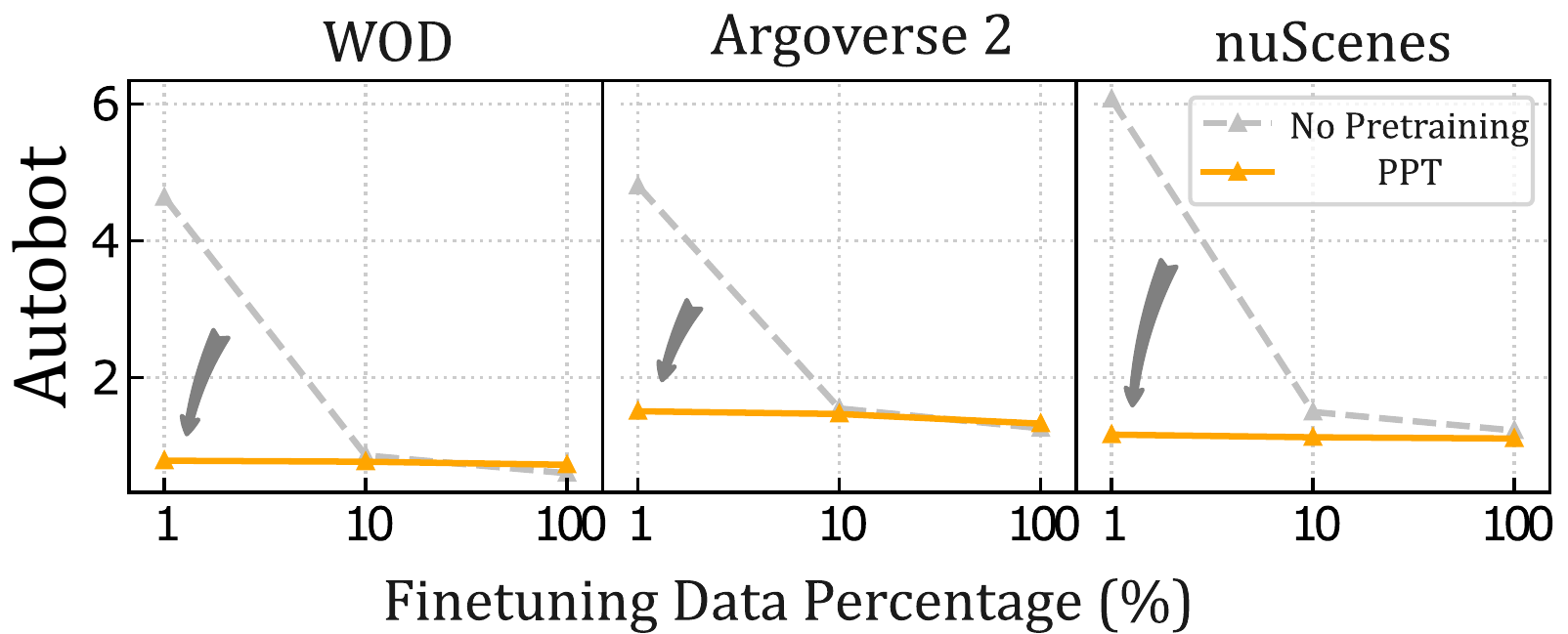} \\
\includegraphics[width=0.8\linewidth]{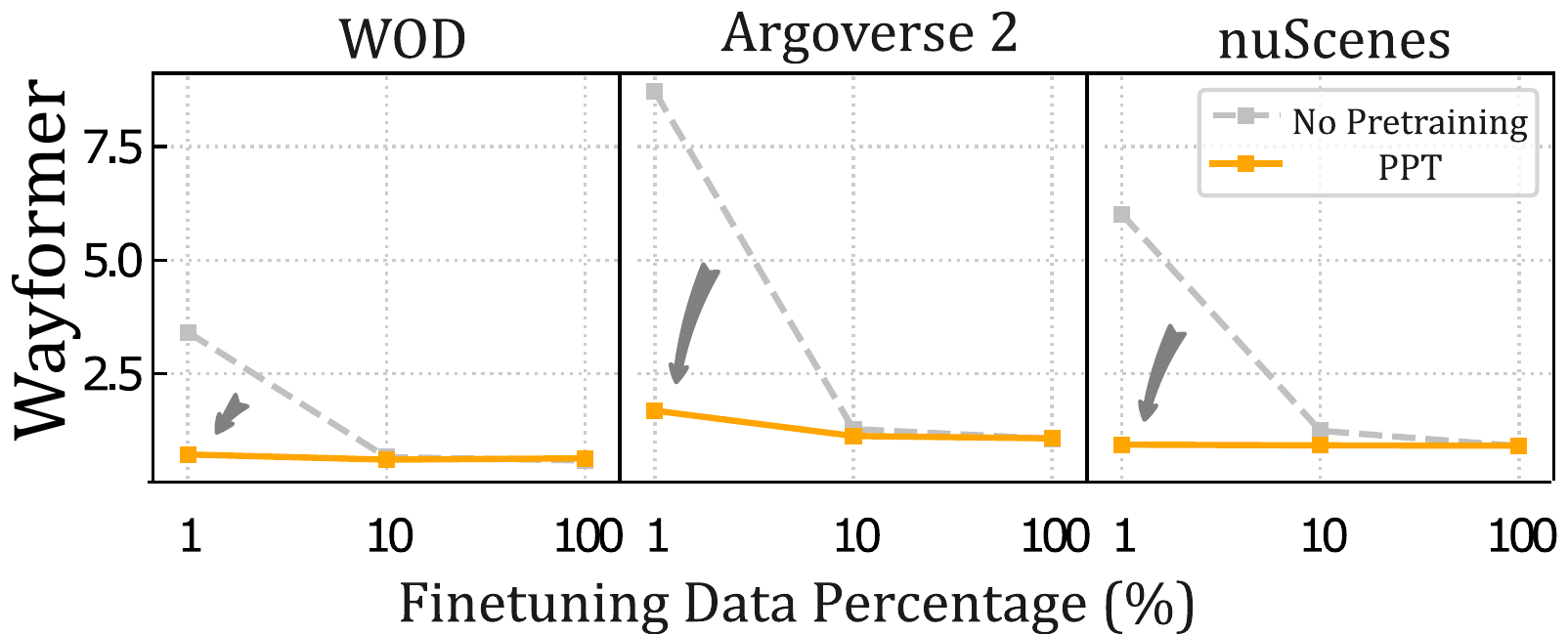} \\
\end{tabular}
}
\caption{\textbf{No pretraining \vs pretraining} with PPT using Autobot \cite{autobot} or Wayformer \cite{Wayformer}.}\label{fig:additional_results_wayformer_autobot}\label{sec:add_res}
\end{figure}

\noindent\textbf{Additional results using PPT.} \xyh{Additionally, we show in \autoref{fig:additional_results_wayformer_autobot}, the results using Autobot~\cite{autobot} and Wayformer~\cite{Wayformer}, respectively. We show that in both methods, PPT generally improves motion forecasting performance compared to "No pretraining", especially in the low-labeled-data regimes.}

\begin{figure}[h]
\centering
\setlength{\tabcolsep}{2pt}
\resizebox{\linewidth}{!}{%
\begin{tabular}{cccc}
\includegraphics[width=0.315\linewidth]{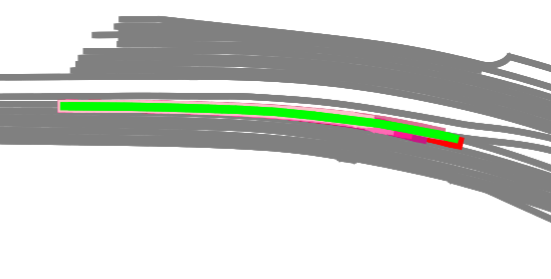}&
\includegraphics[width=0.315\linewidth]{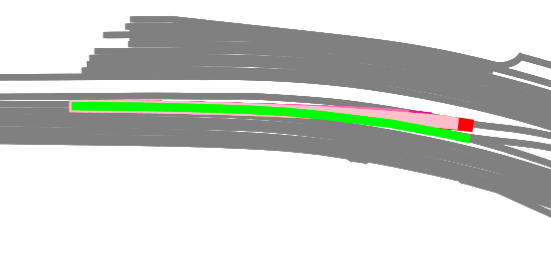}&
    
\includegraphics[width=0.315\linewidth]{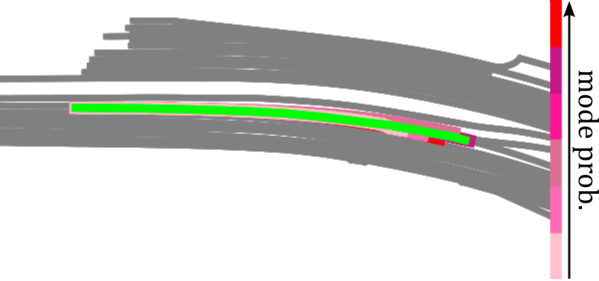} \\

\includegraphics[width=0.315\linewidth]{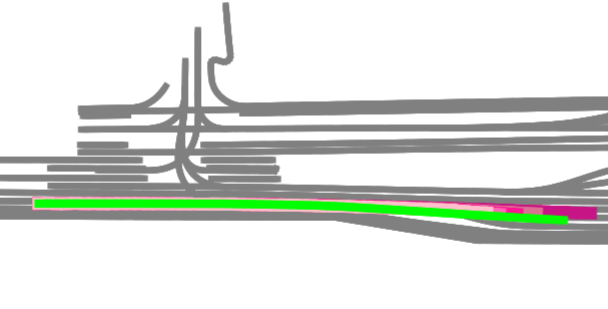} & 

\includegraphics[width=0.315\linewidth]{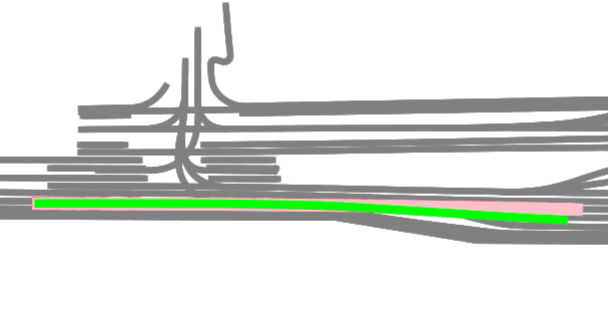} & 

\includegraphics[width=0.315\linewidth]{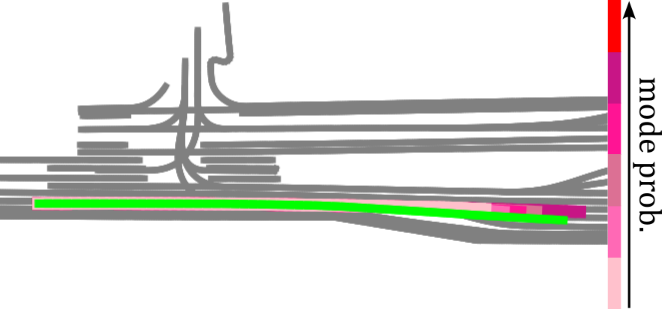} \\
\includegraphics[width=0.315\linewidth]{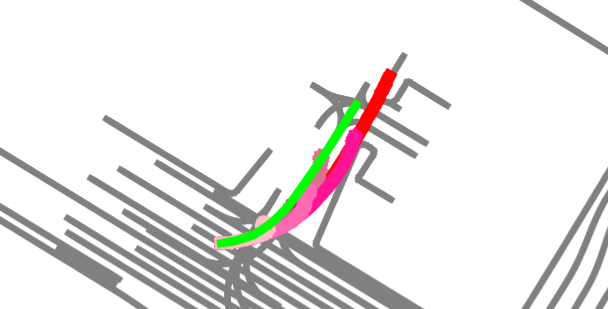}& 
\includegraphics[width=0.315\linewidth]{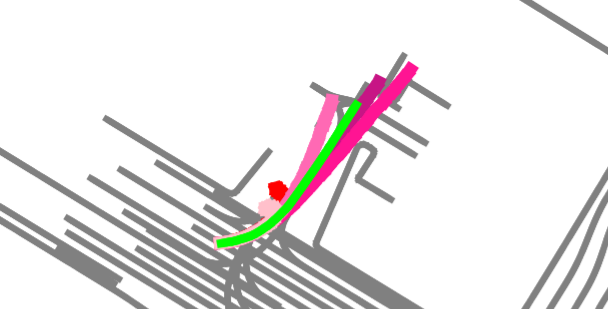} &
\includegraphics[width=0.315\linewidth]{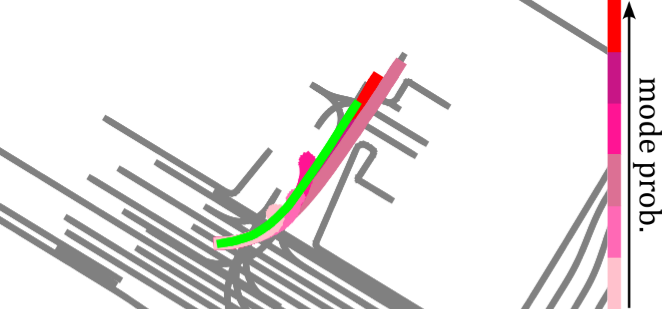} \\
  (a) PPT \textit{w/o} finetuning & 
  (b) No pretraining&
  (c) PPT \textit{w/} finetuning 
\end{tabular}
}
\caption{\textbf{Qualitative results.} We show some qualitative results comparing MTR trained with (a) PPT \textit{w/o} finetuning, (b) No pretraining, and (c) PPT \textit{w/} finetuning. Ground-truth future trajectories are shown in green. Predictions are colored in red with different intensities referring to mode probabilities, \ie the predicted trajectories with the darkest red are predicted at the highest probability.}
\vspace{-6mm}
\label{fig:qualitative}
\end{figure}

\noindent\textbf{Forecasting qualitative results.}
\xyh{We show some qualitative examples in \autoref{fig:qualitative} from models trained with (a) PPT \textit{w/o} finetuning, (b) No pretraining, and  (c) PPT \textit{w/} finetuning. Overall, they perform reasonably well even for the model with only~\method{} pretraining. In the first two rows of \autoref{fig:qualitative}, training with~\method{}, (a) and (c), provides more diverse and precise forecasts (better overlapped with GT), compared to less diverse modes and less precise predictions of (b) w/o~\method{}. In the last row, the trajectory of (b) with the highest probability falls behind the ground-truth prediction in green.
}

\section{Conclusion}

We establish~\method{} as a simple and effective pretraining framework toward robust motion forecasting in diverse driving contexts. Without aiming for single-label clean annotations, \method{} embraces noise and diverse trajectories without human effort, offering significant advantages: (1) \method{} achieves superior results across various datasets, in particular with limited labeled data for finetuning. (2) Improved generalization capabilities of models trained with PPT, as tested with cross-domain, end-to-end, and multi-class settings. Further analyses show the impact of data noise levels, the importance of data variability, and the independence of HD maps and post-processing. (3) \xyh{Finally, by scaling the pretraining pseudo-labels through combining up to 8.6M trajectories, we already observed diminishing returns with more detectors, suggesting that performance likely saturates beyond a certain point.}

\section*{Acknowledgments}
{This work was supported by the ANR grant MultiTrans (ANR-21-CE23-0032). This work was performed using HPC resources from GENCI-IDRIS (Grant 2023-GC011015459). This research
received the support of EXA4MIND project, funded by a European Union’s Horizon Europe Research and Innovation Programme, under Grant Agreement N°101092944. Views and opinions expressed are however those of the author(s) only and do not necessarily reflect those of the European Union
or the European Commission. Neither the European Union nor the granting authority can be held responsible for them. We also acknowledge EuroHPC Joint Undertaking for awarding the project ID EHPC-REG-2024R02-210 access to Karolina, Czech Republic.}

{\small
\bibliographystyle{IEEEtran}
\bibliography{main}
}
\end{document}